\documentclass[journal]{IEEEtran}
\usepackage[scaled=.92]{helvet}
\usepackage{times}
\usepackage{graphicx}
\usepackage{subfigure}
\usepackage{amsmath}
\usepackage{parskip}
\usepackage{algorithm}
\usepackage{algorithmic}
\usepackage{multirow}
\usepackage{color}

\usepackage[labelfont=bf,textfont=it]{caption}

\begin{document}
%
\title{Single Image Brightening via Multi-Scale Exposure Fusion with Hybrid Learning}
%
%
%
\author{Chaobing Zheng \dag, Zhengguo Li \dag, Yi Yang and Shiqian Wu*
		\thanks{\dag Joint first authors. The corresponding author is Shiqian Wu.}
		\thanks{Chaobing Zheng, Yi Yang and Shiqian Wu are with the Institute of Robotics and Intelligent Systems,  School of Information Science and Engineering, Wuhan University of Science and Technology, Wuhan 430081, China(e-mails: zhengchaobing@wust.edu.cn, yangyi@wust.edu.cn, shiqian.wu@wust.edu.cn).}
		\thanks{Zhengguo Li is with the Institute for Infocomm Research, Singapore, 138632, (email: ezgli@i2r.a-star.edu.sg).}}

%
%

\markboth{}
{Shell \MakeLowercase{\textit{et al.}}: Bare Demo of IEEEtran.cls
for Journals}
%



\maketitle

\begin{abstract}
A small ISO and a small exposure time are usually used to capture an image in back- or low-light condition which results in an image with negligible motion blur and small noise but looks dark. In this paper, a single image brightening algorithm is introduced to brighten such an image. The proposed algorithm includes a unique hybrid learning framework to generate two virtual images with large exposure times. The virtual images are first generated via  intensity mapping functions (IMFs) which are computed using camera response functions (CRFs) and this is a model-driven approach. Both the virtual images are then enhanced by using a data-driven approach, i.e. a residual convolutional neural network to approach the ground truth images. The model-driven approach and the data-driven one compensate each other in the proposed hybrid learning framework. The final brightened image is obtained by fusing the original image and two virtual images via a multi-scale exposure fusion algorithm with properly defined weights. Experimental results show that the proposed brightening algorithm outperforms existing algorithms in terms of MEF-SSIM metric. 
\end{abstract}



\IEEEpeerreviewmaketitle

\maketitle

\section{Introduction}
Exposure time and ISO value are the most important factors which can be considered together in various lighting conditions to capture high quality images. For example, in back- or low-light condition, there are two common settings of exposure time and ISO value.  The first one is  a long exposure time and a small ISO value.  The captured image is clean but blurred. The second one is a small exposure time and a large ISO value, and the image is sharp but noisy. Clearly, both settings have difficulty capturing a high quality image, especially when the captured scene is with high dynamic range (HDR). A recommended setting in \cite{CRFs2,1lizg2017} is a small exposure time and a small ISO value. The captured image is clean and sharp, and most of the high-light regions are not over-exposed, but the image is dark. Therefore, a single image brightening algorithm is highly demanded to increase the brightness so as to obtain a clean, sharp and bright image. Four possible challenging problems in single image brightening are: 1) noise in under-exposed regions could be amplified; 2) the high-light regions could be washed out; 3) there could be lightness distortion in the brightened image \cite{LECARM}; and 4) color could be distorted for the pixels in the under-exposed regions \cite{Li}.

There are two types of single image brightening algorithms. One is model-driven image processing technologies \cite{1celebi2015,1fu2015,1park2017,1yue2017,1li2018,1zhang2019,1liu2019,1hao2020} and the other is data-driven methods such as deep learning ones \cite{1chen2018,DeepUPE}. Inputs to a model-driven image brightening algorithm are an image (images) to be processed and the related visual prior(s) \cite{CRFs2}. The prior(s) is (are) achieved by researchers according to their R\&D experience. The model-driven algorithms are built on human's intelligence. No off-line training is required by the model-driven algorithm but the on-line computational cost could be high. Inputs to a data-driven image brightening algorithm are an image (images) to be processed and the corresponding ground truth image (images) \cite{1chen2018,DeepUPE}. Off-line training is required by the data-driven algorithm but the on-line computational cost could be low. Each network is trained for each camera in \cite{1chen2018}. Considering the pros and cons of these two types of methods, fusing a model-driven method and a data-driven one might be an effective way to study single image brightening. Such a framework is called hybrid learning. Two objectives of this paper are: 1) to explore the feasibility of such a framework rather than a sophisticated neural network for deep learning such that the model-driven and data-driven methods can {\it compensate} each other; and 2) to address the four challenging problems in single image brightening.

In this paper, a new hybrid learning framework is introduced for single image brightening under an assumption that the camera response functions (CRFs) are available. This assumption is not an issue if the proposed algorithm is embedded in a digital camera, as different exposure images can be acquired by the camera in advance  to estimate the CRFs via the method in \cite{CRFs}. Same as the algorithm in \cite{CRFs2}, two virtual images are generated to brighten different parts of the captured image. Instead of using the fixed ratio strategy in \cite{CRFs2}, these two images are first generated via a model-driven method, i.e. by using the intensity mapping functions (IMFs) which can be obtained from the CRFs.  As such, there is no lightness distortion in the brightened images. It is noted that noise is amplified and color is distorted in under-exposed regions if the IMFs are used directly \cite{Li}. To handle the color-distorted problem, a fixed ratio strategy is designed to brighten each pixel with at least one under-exposed color channel. The ratio is properly determined by solving a least-square optimization problem to prevent possibly visible seam from appearing at the boundaries among the under-exposed regions and their neighboring regions. Besides reducing the color distortion, it is also necessary to avoid amplifying noise in the under-exposed regions. A simple edge-preserving smoothing method is provided to address the problem. The input image is decomposed into a base layer and a detail layer by using a weighted guided image filter (WGIF) in \cite{WGIF}, and only the base layer is multiplied by the ratio to brighten the underlying image.

Due to limited representation capability of the IMFs, there is visible difference between the virtual images and their ground truth ones. A deep learning method is adopted to enhance the virtual images such that they are closer to their ground truth images. Since there could be color distortion produced by the model-driven method, one more color loss function is introduced to reduce the possible color distortion in the enhanced virtual images. As such, the proposed loss function is composed of restoration loss and color loss. Under-exposed regions in a low-lighting image are usually dominated by the noise, the randomness of noise brings some trouble to train the network. An adaptive weight is thus incorporated in the restoration loss function to  mitigate its influence on the feedback adjustment, making the network easier to converge. Both the convergence speed and the accuracy of the hybrid learning are improved compared to the existing residual deep learning.

Besides increasing the brightness of the input image, it is also important to prevent the brightest regions of the input image from being washed out in the brightened image. Two new weighting functions are proposed to achieved the objective. All the input image and two virtual images are fused via the multi-scale exposure fusion (MEF) algorithm in \cite{TOMP} to produce the final image with the new weighting functions and the weighting functions in \cite{TOMP}.   Experimental results show that the proposed algorithm outperforms other single image brightening algorithms. Same as the network in \cite{1chen2018}, each hybrid learning framework is trained for each camera. In other words, both the proposed framework and the network in \cite{1chen2018} could be embedded in a smart phone or a digital camera. On the other hand, our input is an sRGB image rather than a raw image in \cite{1chen2018}. Besides brightening darkest regions of the sRGB image, brightest regions of the sRGB image are prevented from being washed out while they could be are washed out by the existing brightened algorithms. In summary, our contributions are highlighted as follows:

1) A hybrid learning is formulated by integrating data-driven and model-driven methods for single image brightening. Such solution combines the advantages of both types of methods to generate two virtual images of higher quality. This is a good example to show that the model-driven and data-driven methods can {\it compensate} each other.

2) A new restoration loss function is introduced, in which adaptively weights are assigned to the loss caused by suspicious noise, and the convergence speed is improved.

3) A database, which consists of 300 low, medium and high exposure image triplets, is built up. Only the exposure time is changed while other configurations of cameras are fixed. Both camera shaking and object movement are strictly controlled to ensure that only the illumination is changed.

The rest of this paper is organized as below. Relevant works on single image brightening are reviewed in Section \ref{new}. The proposed hybrid learning framework is summarized in Section \ref{paradigm1}. Two initial virtual images are generated by using the IMF in Section \ref{paradigm2}. They are enhanced via a deep learning method in Section \ref{paradigm3}. The input image and the two virtual images are fused together in Section \ref{fusion} to produce the brightened image. Extensive experimental results are provided in Section \ref{paradigm4} to verify the proposed hybrid learning framework. Finally, conclusion remarks are drawn and future works are discussed in Section \ref{paradigm5}.

\section{Relevant Works on Single Image Brightening}
\label{new}

In this section, existing works on single image brightening are summarized under two categories.

Conventional brightening algorithms such as histogram equalization, gamma correction et al. are simple and intuitive ways to enhance a low-lighting image. Although these methods can stretch the contrast of these images, and tackle the low visibility, other problems arise, such as noise  amplification, detail loss in bright areas \cite{sature}. To address these problems, many single image brightening algorithms have been proposed in the past decade.

The single image brightening algorithms can be classified into two categories. One category is model-driven methods, such as low lighting image enhancement (LIME) \cite{LIME}, which extends the concept of Max-RGB \cite{Retinex} to the pixel level. The illumination of each pixel is first estimated as $\max(R, G, B)$ individually. A structure prior is then imposed on the estimated illumination to obtain the final illumination map. The final image is obtained using the illumination map. Li et al. \cite{CRFs2} proposed an image brightening algorithm using the MEF, in which two virtual images with large exposure times are generated via a fixed ratio strategy. The input image and the two virtual images are fused via the MEF algorithm to produce the final image. The fixed ratio strategy can avoid color distortion in under-exposed regions. However, it assumes that the brightness relationship is linear between two co-located pixels in the two differently exposed images, which could result in lightness distortion  \cite{LECARM} which is one challenging problem in single image brightening. An interesting CRF based strategy was proposed in \cite{LECARM} for the single image brightening. Experimental results show that this method can reduce lightness distortion. The CRF based strategy was ever used in \cite{Li} to study ghost removal for differently exposed images with moving objects. Selecting one of the differently exposed images as the reference image.  All pixels in other images are classified into consistent and inconsistent pixels. All the consistent pixels are kept while all the inconsistent pixels are corrected using the CRFs.  As indicated in \cite{Li}, it is very challenging to correct an inconsistent pixel in an under-exposed region via the CRFs if the exposure time of the reference image is smaller than that of  the image to be corrected. Color could be distorted and noise could be amplified by the CRFs, which are also two challenging problems for single image brightening.

The other category is data-driven methods such as deep learning ones. The deep learning has been widely applied to address image processing problems including single image haze removal \cite{SSIHR,SIDDL,LATP}, single image rain removal \cite{Derain1,Derain2}, single image denoising \cite{denoise1,denoise2}, as well as single image brightening. Li et al. \cite{LightCNN} designed a LightNet to predict the mapping relations between the weakly illuminated image and the corresponding illumination map. It is shown that excellent performance has been achieved if the weakly illuminated image is with good quality but it fails to brighten a weakly illuminated image with low quality such as containing noise or JPEG compression distortion. Wei et al. \cite{Retinex-Net} combined deep learning with a Retinex model to design a Retinex-Net which is composed of a Decom-Net for separating reflectance from illumination and an Enhance-Net for illumination adjustment. Wang et al. \cite{DeepUPE} introduced a new neural network to learn an image-to-illumination mapping rather than an image-to-image mapping. Chen et al. \cite{1chen2018} proposed an interesting deep learning method to map a very dark raw image to a bright image, and each network is trained for each camera in \cite{1chen2018}. They intended to generate perceptually good images in low-light conditions while the brightened image looks blurry and the high-light regions are washed out in the brightened image. The data-driven approach has an advantage to obtain a mapping function without hand-crafted parameter tuning. Nevertheless, such technique requires large amount of training data. All the deep learning based brightening algorithms focused on increasing the brightness of the input image. Unfortunately, the high-light regions of the input image could be washed out. It is necessary to address this challenging problem for single image brightening.

Due to the pros and cons of these two different categories of methods, it is worth fusing a model-driven method and a data-driven one for single image brightening. The objectives of this paper are to explore such a fusion of the model-driven method and the deep learning method so that they can {\it compensate} each other, and to address the four challenging problems for single image brightening.

\section{The Proposed Brightening Algorithm}
\label{paradigm1}

In this section, a single image brightening algorithm is proposed by introducing a hybrid learning framework which is a combination of model-driven and data-driven methods. Same as \cite{1chen2018}, each hybrid learning framework is trained for each camera. The brightness of the input image is increased while the high-light regions are prevented being washed out.

\begin{figure*}[htb]
	\centering
	\includegraphics[width=0.9\textwidth]{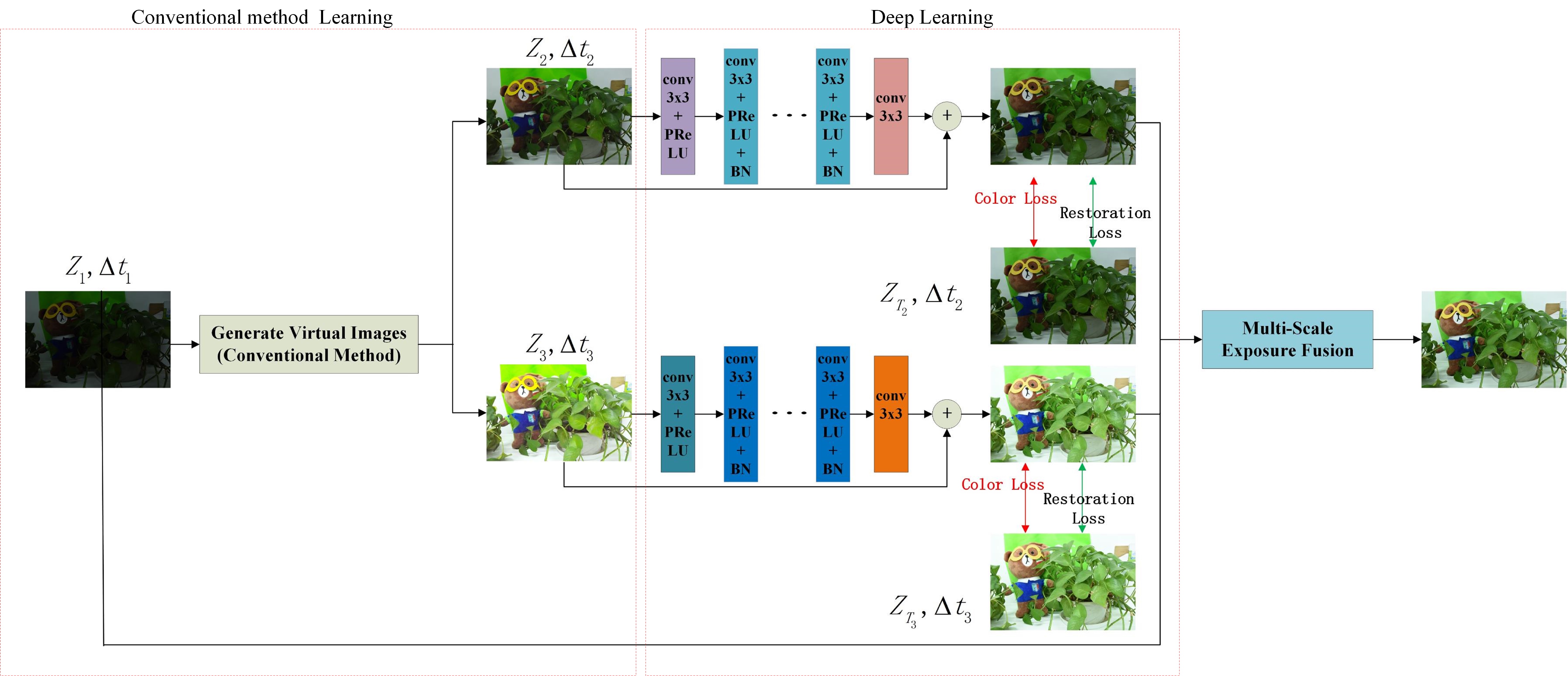}
	\caption{The diagram of the proposed single image brightening algorithm via the MEF. Two virtual images $Z_2$ and $Z_3$ with ${\Delta} t_2$ and ${\Delta} t_3$, which are larger than ${\Delta} t_1$, are first generated by a model-driven method, i.e. the IMF based method.   $({Z_{T_2}}-{Z_2})$ and $({Z_{T_3}}-{Z_3})$ are then learnt via the data-driven residual CNN to enhance the initial virtual images. The input image and the two virtual images are finally fused to obtain a brightened image. The brightness of the image is increased while the brightest regions are prevented from being washed out.}
	\label{Fig_two}
\end{figure*}

Let $Z_1$ be an eight-bit image which is captured at back- or low-light condition. Two eight-bit virtual images $Z_2$ and $Z_3$ with large exposure times are produced by using a model-driven method. The ground truth images of $Z_2$, $Z_3$ are denoted as $Z_{T_2}$ and $Z_{T_3}$, respectively. They are captured together with the image $Z_1$ by using the method in \cite{CRFs}. Fig. \ref{Fig_two} summarizes the pipeline of our network for a single image brightening via multi-scale exposure fusion with hybrid learning. Since one objectives of this paper is to explore the feasibility of fusing model-driven and data-driven methods rather than a sophisticated neural network for deep learning, the CNN used in the proposed framework is on top of the network in \cite{DNCNN}  while the ReLU is replaced by the PReLU.

 The key component of the proposed algorithm is to generate two high-quality virtual images using a new concept of hybrid learning, which combines a model-driven method and a deep data-driven one. The necessity on such a fusion can be elaborated by borrowing wisdom from the field of nonlinear control systems \cite{Ninlinear,11li2005}, namely, modelled dynamics and unmodelled dynamics are two concepts in the field of nonlinear control systems \cite{Ninlinear}.  $Z_2$ and $Z_3$ produced by using the IMFs are the modeled information of $Z_{T_2}$ and $Z_{T_3}$ while $(Z_{T_2}-Z_2)$ and $(Z_{T_3}-Z_3)$ are the unmodelled information. The modelled information and the unmodelled information are analogous to the modelled dynamics and the unmodelled dynamics.

 Different from existing data-driven methods and model-driven ones, the virtual images ${Z_2}$ and ${Z_3}$ are obtained by using a model-driven method in advance, then the unmodelled information $({Z_{T_2}}-{Z_2})$ and $({Z_{T_3}}-{Z_3})$ are learned by using a data-driven deep residual convolutional neural network. Clearly, the quality of the initial  virtual images can be improved if part of the unmodelled information can be further represented. Such a framework can be regarded as hybrid learning.

The proposed hybrid learning has the following advantages: Firstly, compared with the model-driven method, the virtual images produced by the proposed method is enhanced by compensating unmodelled information. Secondly, compared with the data-driven solution,  the hybrid learning converges fast and requires fewer training samples, because $({Z_{T_2}}-{Z_2})$ and $({Z_{T_3}}-{Z_3})$ are sparser than ${Z_2}$ and ${Z_3}$. Thus, it is easy to train the latter neural network using a residual network \cite{HKM}. Clearly, the model-driven and data-driven methods can {\it compensate} each other.

Finally, the input image $Z_1$ and two virtual images $Z_2$ and $Z_3$ are fused together using the MEF algorithm in \cite{TOMP} to produce the final image. The details on the proposed algorithm are given in the subsequent sections.

\section{Generation of Initial Virtual Images}
\label{paradigm2}

Two initial virtual images  $Z_2$ and $Z_3$ that are brighter than the input $Z_1$ will be produced by using the IMFs in this section.

Let the CRF be ${f_l}(\cdot)$, and the exposure times of  ${Z_1}$, ${Z_2}$ and ${Z_3}$ be ${{\Delta}t_1}$, ${{\Delta}t_2}$ and ${{\Delta}t_3}$, respectively. Without loss of generality, it is assumed that  ${{\Delta}t_3} > {{\Delta}t_2} > {{\Delta}t_1}$. It has been shown in \cite{Y} that there is relative brightness change in the fused image if the exposure ratios are too large. Thus, $\Delta t_3$ and $\Delta t_2$ are selected as $16\Delta t_1$ and $4\Delta t_1$ in this study.
The IMF between input image and two virtual images can be expressed as follows:
\begin{equation}
\label{IMFS}
{\Lambda}_{1,i,l}(z)={f_l}(\frac{{f^{-1}_l}(z){{\Delta}t_i}}{{\Delta}t_1})\; ;\; i=2, 3,
\end{equation}
where $l$ is a color channel. ${f^{-1}_l}(\cdot)$ is the inverse function of the CRF ${f_l}(\cdot)$.

Instead of using the fixed ratio strategy in \cite{CRFs2}, two initial virtual images are generated by using the IMFs. The lightness distortion is prevented from appearing in the brightened images. As mentioned in \cite{Li}, if the pixel value $z$ is larger than a threshold ${\xi}_L$,  ${\Lambda}_{1,i,l}(z) $ is a one-to-one mapping, which is reliable. Otherwise, it is not reliable due to a one-to-many mapping. Here, the threshold ${\xi}_L$ is determined by the quality of the camera. Both cases are considered to produce the two initial virtual images as follows:

{\it Case 1}: The $l$th color channel of pixel $p$, ${Z_{1,l}(p)} $ is bigger than the threshold ${\xi}_L$ for each channel $l$. The pixel values corresponding to the two virtual images can be computed by using the IMF. The virtual pixels $Z_2(p)$ and $Z_3(p)$ are computed as
\begin{align}
Z_2(p)&=[{\Lambda}_{1,2,1}({Z_{1,1}(p)}),{\Lambda}_{1,2,2}({Z_{1,2}(p)}),{\Lambda}_{1,2,3}({Z_{1,3}(p)})],\\ Z_3(p)&=[{\Lambda}_{1,3,1}({Z_{1,1}(p)}),{\Lambda}_{1,3,2}({Z_{1,2}(p)}),{\Lambda}_{1,3,3}({Z_{1,3}(p)})].
\end{align}

{\it Case 2}: ${Z_{1,l}(p)} $ is smaller than the  threshold ${\xi}_L$ for at least one channel, and the IMF is not reliable, which yields color distortion. Hence, the fixed ratio strategy in \cite{CRFs2} is adopted to generate the corresponding virtual pixels. Two challenging problems to be addressed are: 1) the fixed ratio strategy amplifies noise in the under-exposed regions of $Z_1$; and 2) there are visible seams at the boundaries among the underexposed regions and their neighboring regions if the ratio is selected as in \cite{CRFs2}. To address the first problem, $Z_1$ is decomposed as a base layer $Z_{1}^{b}$ and a detail layer $Z_{1}^{e}$ by the WGIF \cite{WGIF}.  To address the second problem, the virtual pixels $Z_i(p)(i=2,3)$ are computed as
\begin{equation}
\label{decomposed}
Z_i(p)= \tilde{\gamma}_i  Z_{1}^{b}(p)+Z_{1}^{e}(p),
\end{equation}
where the values of  ${{\tilde{\gamma}}_i}$(i=2,3) are obtained by minimizing the following function:
\begin{equation}
\label{mini}
{\sum_{l=1}^3}{\tilde{w}(Z_{1,l}(p))(\Lambda_{1,i,l}(Z_{1,l}(p))-Z_{1,l}^{e}(p)-{{\tilde{\gamma}}_i}{Z_{1,l}^{b}(p)})^2},
\end{equation}
and the function $\tilde{w}$ is defined as \cite{1yao2012}:
\begin{eqnarray}
&&\hspace{-7mm}\tilde{w}(z) = \left\{ {\begin{array}{*{20}{l}}
	{0;}&{if~}{{\rm{0}} { \le z < {\xi _L}}}\\
	{128 - 3{h_1}^2(z) + 2{h_1}^3(z);}&{if~}{{\xi _L} \le z < {\xi _U}}\\
	{128;}&{otherwise}{}.
	\end{array}} \right.,
\end{eqnarray}
The function ${h_1}(z)$ is given as
\begin{equation}
\label{h}
{h_1}(z)=\frac{\xi _U-z}{\xi _U-\xi _L}.
\end{equation}
Similar to $\xi _L$, the value of $\xi _U$ is related to camera quality. The higher the quality of the camera, the larger the value of $\xi _U$. In this study, we use $\xi _L=5$ and $\xi _U=60$  as the default settings.

\begin{figure*}[htb]
	\centering
	\includegraphics[width=0.9\textwidth]{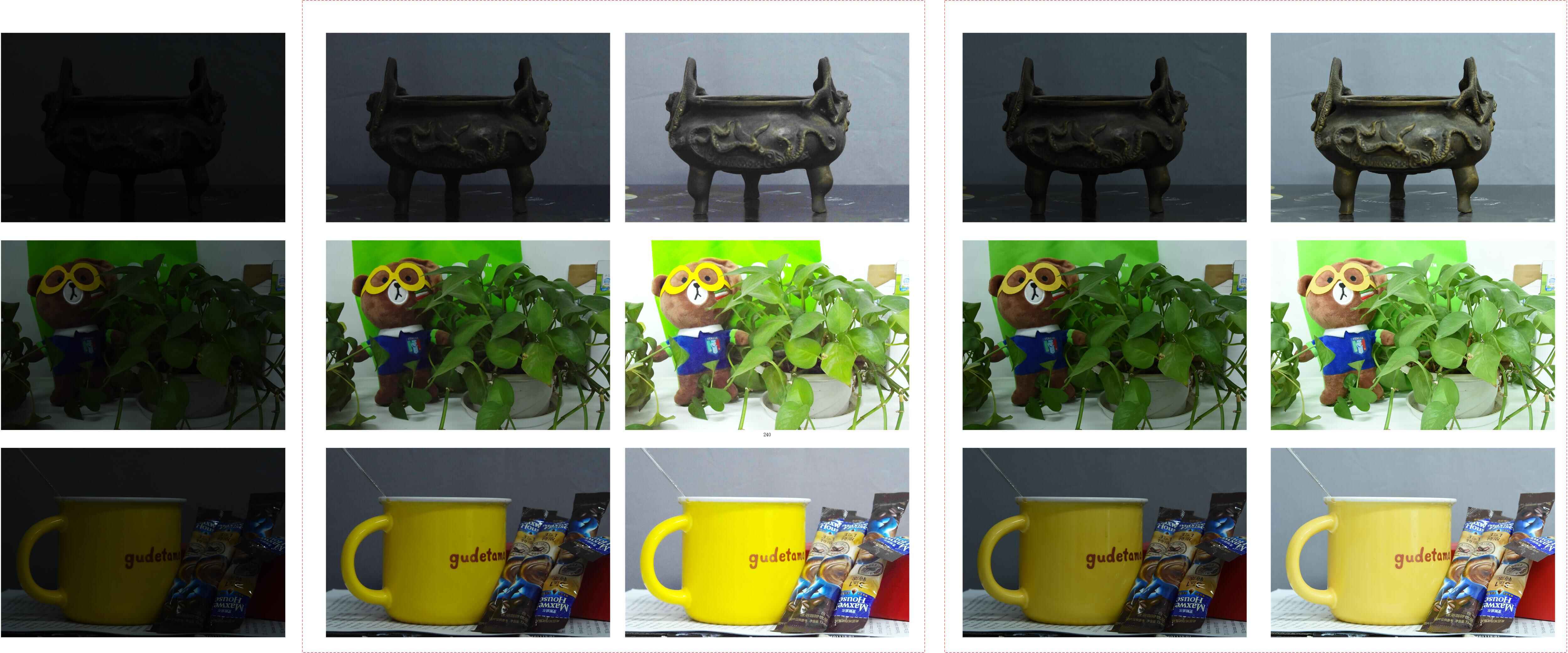}
	\caption{The first column includes three input low-light images which are taken with a Nikon 7200 camera. The ISO value is set as  800, and the exposure time ${\Delta} t_1$ is very short. The second column is the set of initial virtual images $Z_2$'s, the exposure time is ${\Delta} t_2$.  The third column is the set of initial virtual images $Z_3$'s, whose exposure time is ${\Delta} t_3$,  ${\Delta} t_1 < {\Delta} t_2 < {\Delta} t_3$. The fourth column is the set of ground truth images $Z_{T_2}$ in exposure time  ${\Delta} t_2$. The fifth  column is the set of ground truth images $Z_{T_3}$ in exposure time ${\Delta} t_3$.  }
	\label{Fig_one}
\end{figure*}

The resultant virtual images are shown in Fig. \ref{Fig_one}. The virtual images are close to the corresponding ground truth images but the color needs to be corrected. In addition, due to limited representation capability of IMFs, the images $Z_2$ and $Z_3$ do not contain all the information in the ground truth images. Thus, both intensity and color need to be adjusted.

\section{Enhancement of Virtual Images Via Deep Learning}
\label{paradigm3}

\begin{figure*}[htb]
	\centering
	\includegraphics[width=0.9\textwidth]{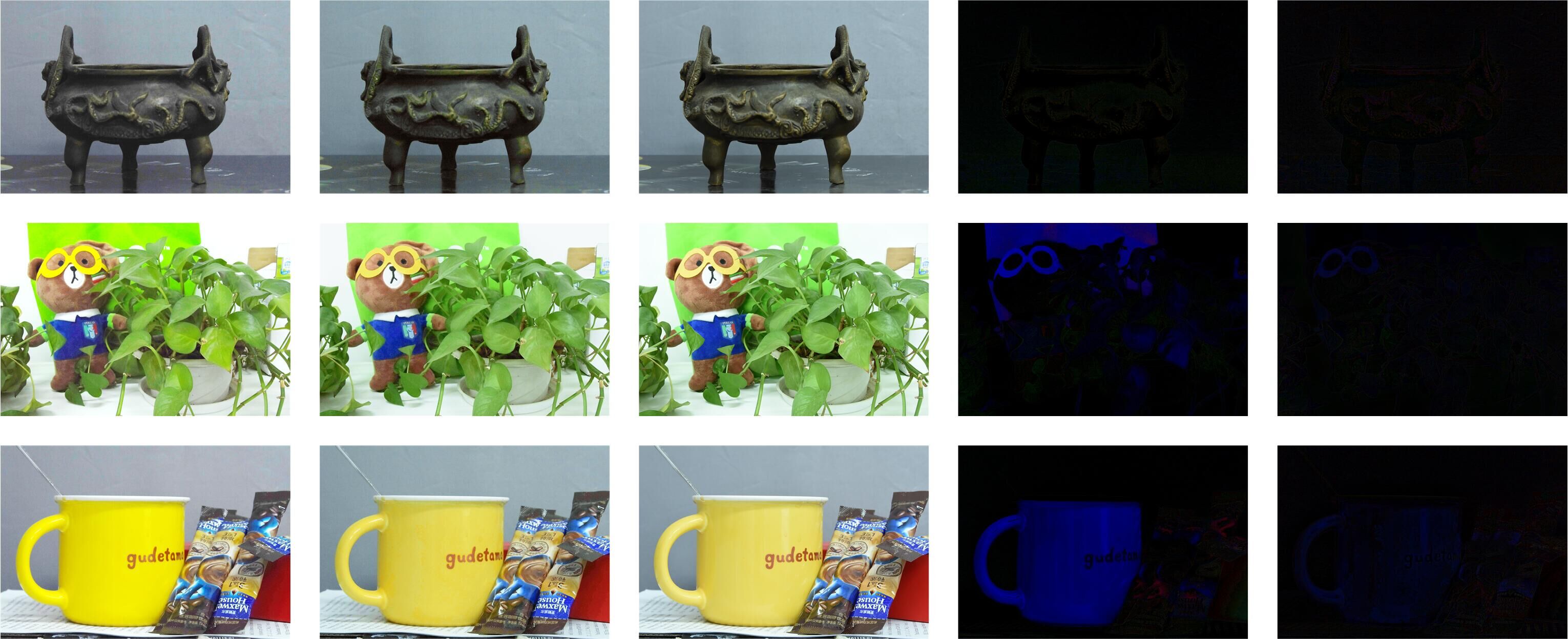}
	\caption{The first column shows the virtual images $Z_3$ generated by the model-driven method, the second column illustrates the results of using the deep learning method to enhance the virtual images $(f(Z_1)+Z_3)$, the third column shows the ground truth images, $Z_{T_3}$, the fourth demonstrates the results of  $(Z_{T_3}-Z_3)$, the fifth shows the results of $(Z_{T_3} - f(Z_1)-Z_3)$.}
	\label{Fig_three}
\end{figure*}

The IMF can be regarded as a model to represent the correlation among different exposed images but its representation capability is limited. As mentioned in the Section \ref{paradigm1}, the unmodelled information $({Z_{T_2}}-{Z_2})$ and $({Z_{T_3}}-{Z_3})$ can be further represented by a deep learning method. The unmodelled information is usually sparse, i.e.,  most values are likely to be zero or small as shown in Fig. \ref{Fig_three}. It is expected that the unmodelled information can be {\it compensated} by a deep neural network.
Similarly, it is challenging for the deep neural network to compensate the IMF in the under-exposed regions due to existence of noise in the regions.

The loss function used in the proposed hybrid learning is defined as
\begin{equation}
L= L_r+w_cL_c,
\end{equation}
where $w_c$ is a constant, which is selected as 2 in this study.

The restoration loss $L_r$ is usually defined as
\begin{equation}
L_r = \sum_{p,l}[Z_{T_i,l}(p)-Z_{i,l}(p)-{f_i}(Z_{1,l}(p))]^2.
\end{equation}
The above loss function can be replaced by the weighted MSE in \cite{1tan2013} which is equivalent to the SSIM but is differentiable. Since the under-exposed regions in $Z_1$ contain random noise \cite{CRFs2}, a content adaptive weight is introduced to the restoration loss so as to reduce the effect of noise on the adjustment parameters. The loss function $L_r$ is given as
\begin{equation}
\label{eq15}
L_r = \sum_{p,l}W_{i,l}(p) [Z_{T_i,l}(p)-{f_i}(Z_{1,l}(p))-Z_{i,l}(p)]^2,
\end{equation}
where the weight function $W_{i,l}(p)$ is expressed as:
\begin{eqnarray}
&&\hspace{-7mm}{W_{i,l}}(z) = \left\{ {\begin{array}{*{20}{l}}
	{1;}&{if~} Z_{i,l}(p)\geq \nu\\
	{\frac{1}{\nu-{Z_{i,l}(p)}};}&{otherwise}{}.
	\end{array}} \right.
\end{eqnarray}
and $\nu$ is a small positive constant and it is empirically selected as 6.0 in this paper if not specified. When the pixel value in position $p$ is smaller than $v$, it may be noise, so a small weight is assigned to the loss.

To minimize the possible color distortion in the two brightened images, the color loss is defined as
\begin{equation}
\label{a1}
L_c=\sum_p\angle (Z_{T_i}(p),  Z_i(p)+{f_i}(Z_{1}(p))),
\end{equation}
where $\angle(Z_{T_i}(p),{f_i}(Z_1(p))+Z_i(p))$ is the angle between two 3D $(R, G, B)$ vectors $Z_{T_i}(p)$ and $ ({f_i}(Z_1(p))+Z_i(p))$. Since the $L_r$ metric only measures the color difference numerically, it cannot ensure that the color vectors have the same direction \cite{DeepUPE}. By employing the color loss in Eq. (\ref{a1}), the possible color distortion is reduced.

The enhanced virtual images is expressed as $({f_i}(Z_1)+Z_i)( i=2,3)$. It is shown in Fig. \ref{Fig_three} that the virtual images enhanced by the deep learning method are much closer to the ground truth images, which implies the unmodelled information is reduced significantly. The color of the plants in the second row is also distorted, while our result looks more natural and much closer to the ground truth. The color of the cup  in the third row is obviously distorted by the model-driven method. The results of our  method are much closer to the ground truth. It is worth noting that the deep learning method works on TensorFlow, and is trained with a mini-batch size of 32. An Adam optimizer with a fixed learning rate of $10^{-4}$ is used to optimize the entire network. Mirroring and cropping are employed to augment training data.

\section{Multi-Scale Fusion of Input Image and Two Virtual Images}
\label{fusion}
The input image $Z_1$ and the two virtual images $Z_2$ and $Z_3$ are fused together to produce the final image. As shown in  \cite{TOMP}, the weighting maps of the differently exposed images play an important role in the MEF algorithm. This section focuses on  defining the weighting maps for the three images while the MEF is the same as \cite{TOMP}.

Besides increasing the brightness of the image $Z_1$, the brightest regions of the image $Z_1$ are prevented from being washed out. To achieve this objective, two different functions are adopted to define the weights for the three images.

One function is used to determine amplification factors of all the pixels in the image $Z_1$. The weight is defined as
\begin{align}
\psi_1(z)=\left\{\begin{array}{ll}
2; &\mbox{if~}z>160\\
1+h_2^2(z)(3-2h_2(z)); &\mbox{if~}160\geq z>128\\
1; &\mbox{otherwise}\\
\end{array}
\right.,
\end{align}
where the function $h_2(z)$ is $(z-128)/32$. The function $\psi_1(z)$ is inspired by a weighting function in \cite{1yao2012}.

The other is to measure contrast, well exposedness level, and color saturation of each pixel in the three images and it is defined the same as in \cite{TOMP}
 \begin{align}
 \psi_2(Z_i(p)) = w_c(Z_i(p))\times w_s(Z_i(p))\times w_e(Z_i(p)),
  \end{align}
  where $w_c(Z_i(p))$, $w_s(Z_i(p))$ and $w_e(Z_i(p))$ measure contrast, color saturation, and well-exposedness of pixel $Z_i(p)$, respectively.

Let $Y_1$ be the luminance component of the image $Z_1$ in YUV color space. The weight of the pixel $Z_1(p)$ is given as
\begin{align}
\label{weight1}
W(Z_1(p)) = \psi_1(Y_1(p))\psi_2(Z_1(p)),
\end{align}
and the weight of the pixel $Z_i(p)(i = 2,3)$ is given as
\begin{align}
\label{weight2}
W(Z_i(p)) = \psi_2(Z_i(p)).
\end{align}

With the above weighting maps, all the images $Z_i(i=1,2,3)$ are fused together via  the existing MEF algorithm  in \cite{TOMP}.
It can be shown in the Eqs. (\ref{weight1}) and (\ref{weight2}) that the pixels in the brightest regions of image $Z_1$ dominate the fusion. As such, the brightest regions of image $Z_1$ are prevented from being washed out by the proposed algorithm.

\section{EXPERIMENTAL RESULTS}
\label{paradigm4}
Extensive experiments are carried out in this section to demonstrate the rationality and effectiveness of the proposed hybrid learning framework. The emphasis is to show how the model-driven method and the data-driven one {\it compensate} each other. Readers are invited to view to the electronic version of the full-size figures and zoom in these figures in order to better appreciate the differences among images. Code is available at \cite{data}.

\begin{figure}[htb]
	\centering
	\includegraphics[width=0.45\textwidth]{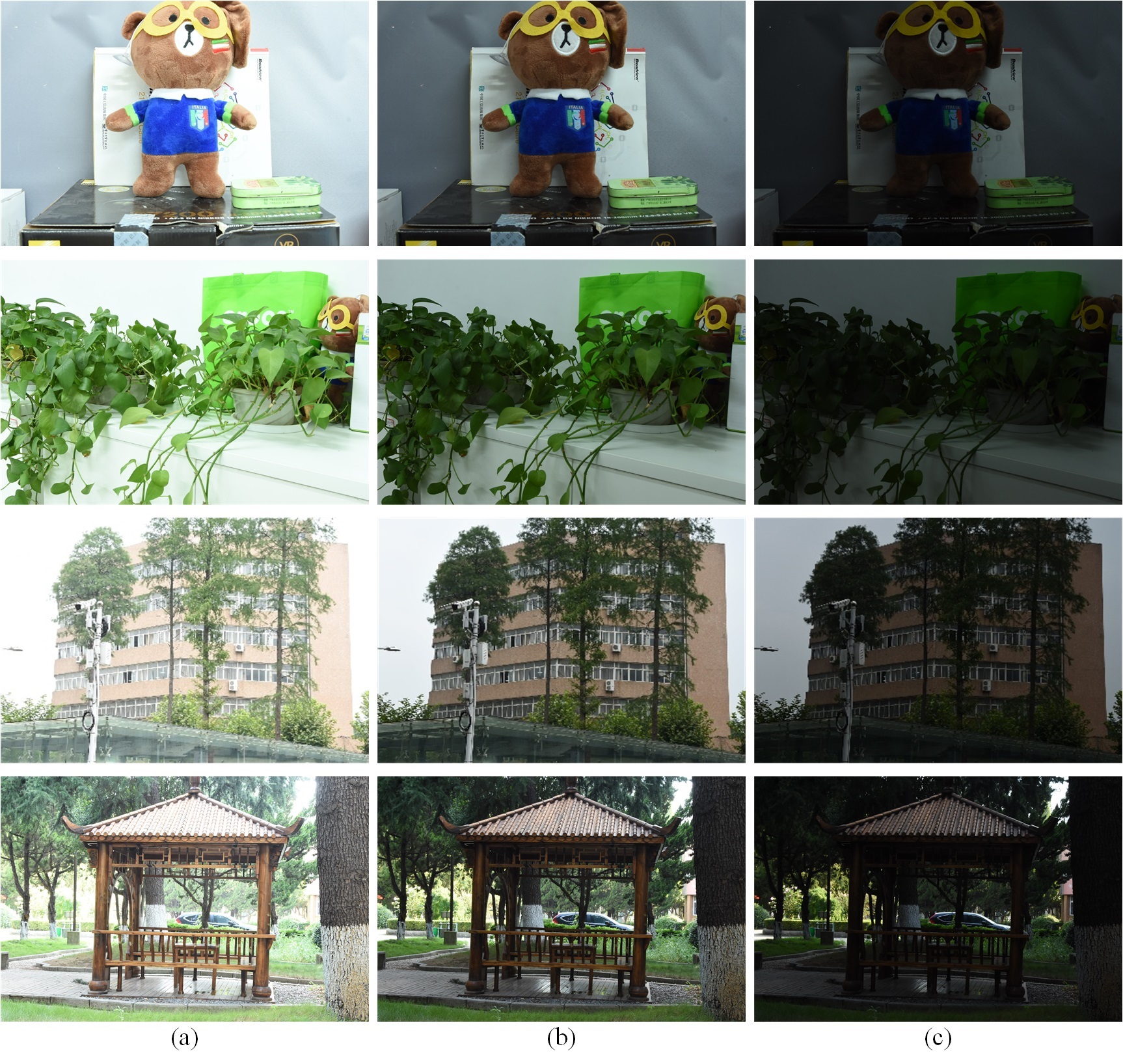}
	\caption{ (a) are high exposure images. (b) are middle exposure images. (c) are low exposure images. The images are collected by changing exposure time, while other configurations of camera are fixed. The camera is fixed to mitigate the effects of jitter, and no moving objects can appear in the image, ensuring that the only variable is illumination. }
	\label{Fig_4}
\end{figure}
\subsection{Dataset}

 We have built up a dataset which comprises 300 sets of differently exposed image. Each set contains three images. The images from difference scenes are captured using Nikon 7200. The ISO is set as  800.  According to \cite{CRFs}, exposure times are different while other configurations of the cameras are fixed. The interval of exposure ratio between them is 2 exposure values (EVs). Both camera shaking and object movement are strictly controlled to ensure that only the exposure time is different, some pictures are shown in Fig. \ref{Fig_4}.

\begin{figure}[t]
	\begin{center}
		\includegraphics[width=0.9\linewidth]{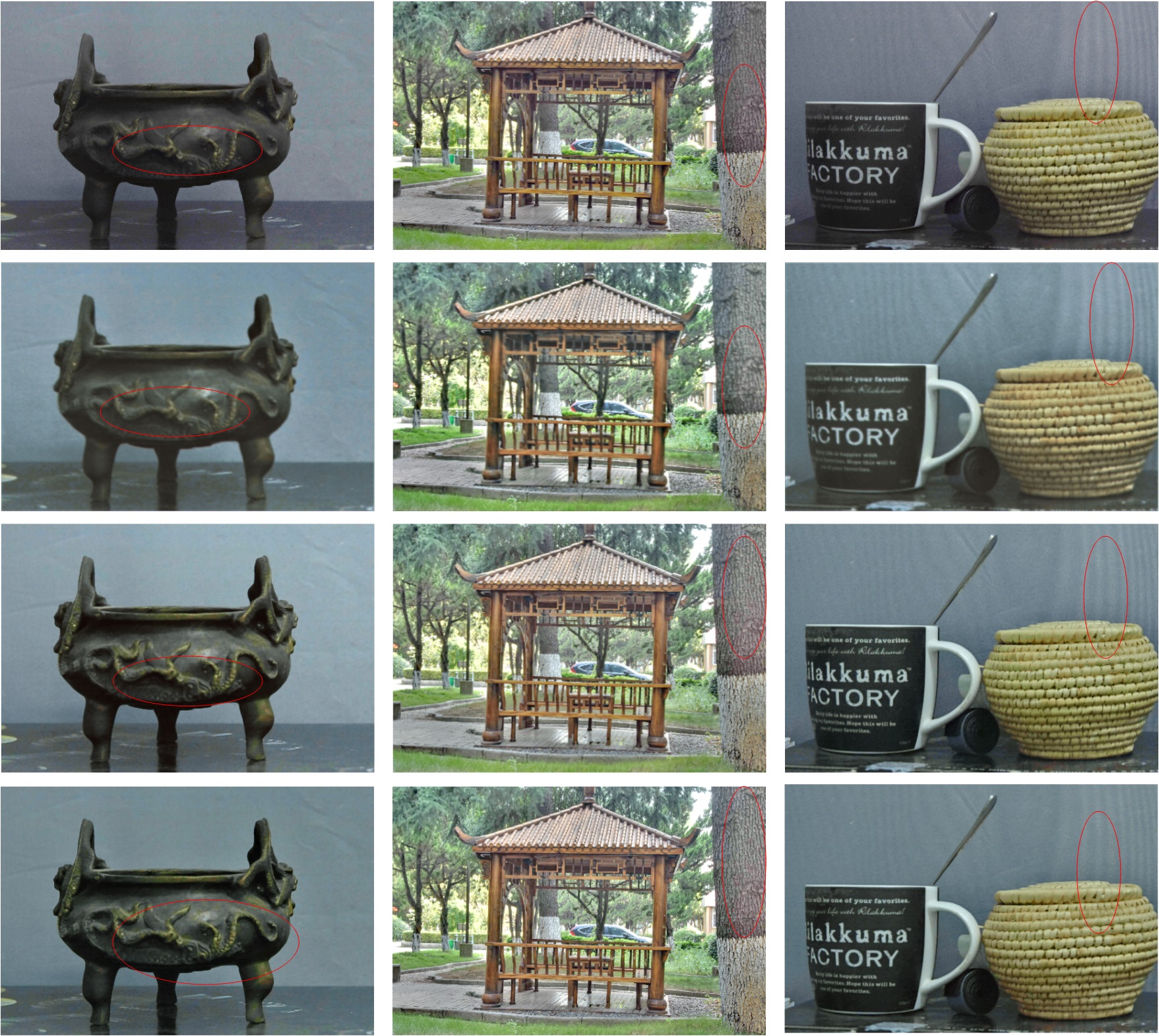}
	\end{center}
	\caption{The first row shows the brightened images by the model-driven method. The second row illustrates the brightened images by the deep learning method. The third row shows the brightened images of the proposed hybrid learning without color loss function. The fourth row demonstrates the brightened images of the proposed hybrid learning.}
	\label{Fig_five1}
\end{figure}

\begin{figure*}[htb]
	\centering
	\subfigure{
		\begin{minipage}[b]{0.45\linewidth}
			\includegraphics[width=1\linewidth]{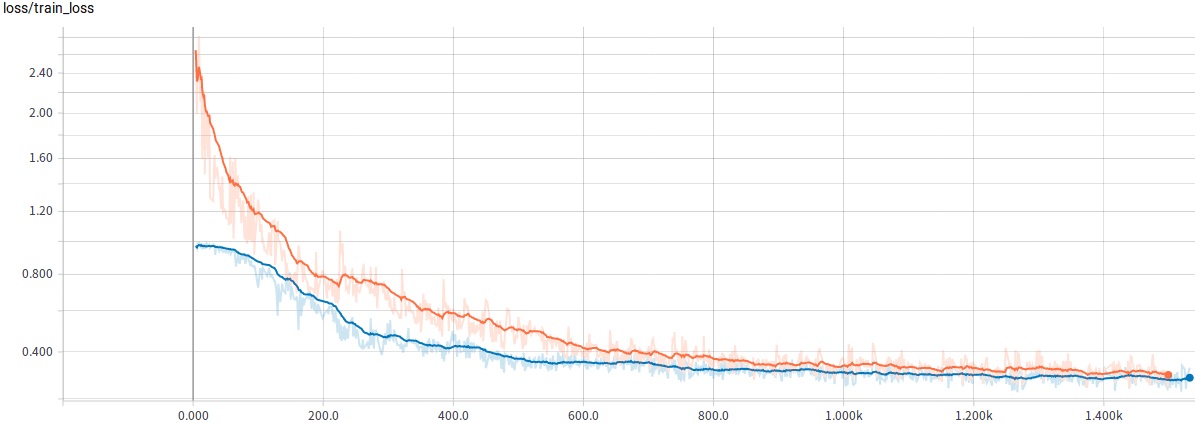}	
			\centerline{(a)}	
	\end{minipage}}
	\subfigure{
		\begin{minipage}[b]{0.45\linewidth}
			\includegraphics[width=1\linewidth]{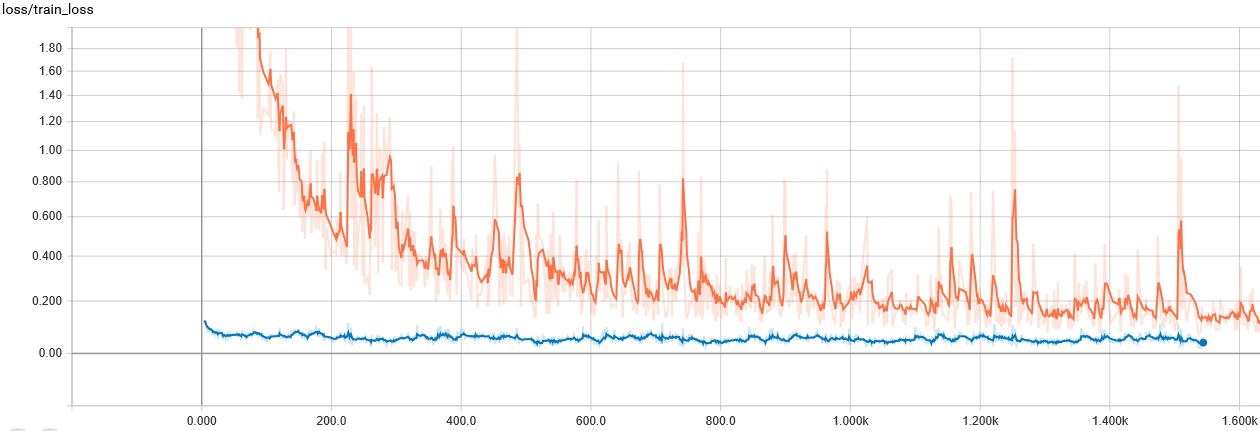}
			\centerline{(b)}
	\end{minipage}}
	\caption{(a) Training curve for intermediate exposure images; (b) Training curve for high exposure images; The Y-axis represents the loss and the X-axis represents the number of iterations. The blue line indicates the results by the proposed hybrid learning, and the red line is the results by the deep learning.}
	\label{fig23}
\end{figure*}

\begin{figure}[t]
	\begin{center}
		\includegraphics[width=0.9\linewidth]{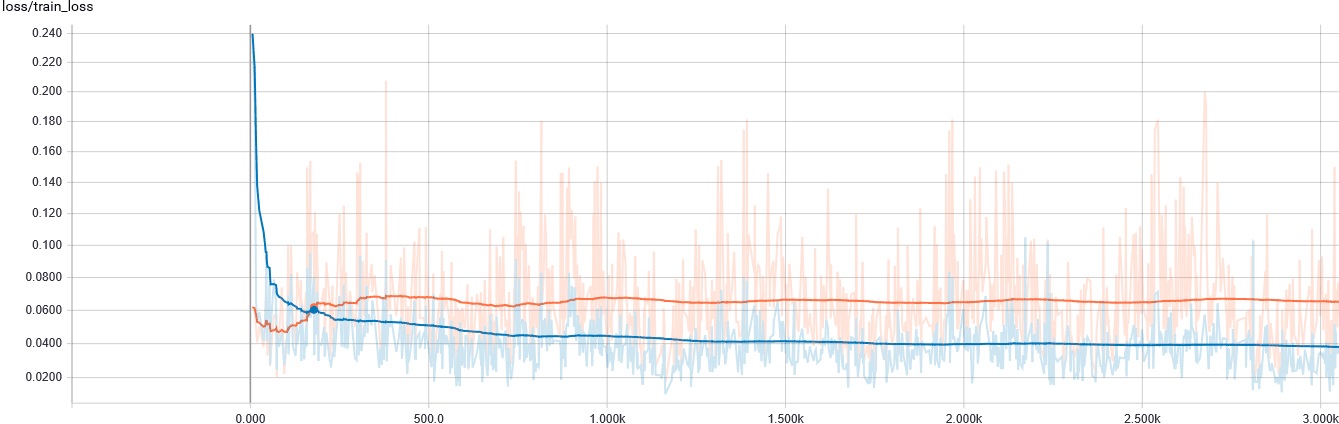}
	\end{center}
	\caption{ Ablation study  on the BN. The Y-axis represents the loss and the X-axis represents the number of iterations. The red line indicates the results without BN, and the blue line is the results with the BN. Clearly, the BN is helpful for the  convergence.}
	\label{Fig_f}
\end{figure}

\subsection{Comparison of Three Different Methods}
 The MEF-SSIM in \cite{MEFSSIM} is adopted to illustrate the performance of three different methods: model-driven method, data-driven, and the proposed hybrid learning. The experimental results are given in Fig. \ref{Fig_five1} and Table \ref{tab1}. Clearly, the proposed hybrid learning can improve both the MEF-SSIM and the visual quality of virtual images. Particularly, the resultant images by the deep learning are blurry as shown in Fig. \ref{Fig_five1}, details are also lost. Sometimes, the results by the deep learning are worse than those by the model-driven method. Clearly, the proposed hybrid learning can make the virtual images much closer to the ground truth images.

\begin{table*}[htb]
	\begin{center}
		\centering
		\caption{MEF-SSIM of Four Different Methods}
		\tabcolsep8pt\begin{tabular}{|c|c|c|c|c|c|}
			\hline
			image& model-driven  & data-driven  & proposed without $BN$      & proposed without $L_c$& proposed   \\\hline
			furnace    & 0.8450     & 0.8494       & 0.8747                           &  0.9185      & \textbf{0.9218}\\
			tea        & 0.8910     & 0.8654       & 0.8935                           &  0.9251      & \textbf{0.9256}\\
			grass      & 0.9611     & 0.9360       & 0.9544    &  0.9686      & \textbf{0.9694}\\
			pavilion   & 0.9565     & 0.9428       & 0.9594    &  0.9605      & \textbf{0.9672}\\
			flower     & 0.9622     & 0.9597       & 0.9612    &  0.9662      & \textbf{0.9665}\\
			cup        & 0.8442     & 0.8421       & 0.8686    &  0.8927      & \textbf{0.8965}\\
			flower pot & 0.8196     & 0.7582       & 0.8319    &  0.8621      & \textbf{0.8660}\\
			succulents & 0.8741     & 0.8692       & 0.8816    &  0.9142      & \textbf{0.9163}\\
			electric fan & 0.9516   & 0.9339       & 0.9528    &  0.9665      & \textbf{0.9668}\\
			bag        & 0.9608     &  0.9301      & 0.9494    &  0.9698      & \textbf{0.9701}\\
			dog        & 0.9070     & 0.8765       & 0.9106    &  0.9329      & \textbf{0.9340}\\
			building   & 0.9369     & 0.9371       & 0.9372    &  \textbf{0.9413}      & 0.9411\\\hline
			average    & 0.9091     & 0.8917       & 0.9146    & 0.9349       & \textbf{0.9368}\\			\hline
		\end{tabular}
		\label{tab1}
	\end{center}
\end{table*}

Besides the quality of virtual images, the convergence speed is also analyzed. There is large difference between the input low-lighting image $Z_1$ and the ground truth medium/high exposure images $Z_{T_i}$. On the other hand, the virtual images $Z_i$ which are generated by the model-driven method  are already close to $Z_{T_i}$. It is easy for the residual CNNs to represent $(Z_{T_i} - Z_i)$, and the networks can converge much faster as shown in Fig. \ref{fig23}. Clearly, the model-driven method and the data-driven one indeed {\it compensate} each other in the proposed hybrid learning framework.

It should be pointed out that the on-line cost of the proposed hybrid learning framework is slighter higher than that of the pure deep learning method due to the inclusion of the data-driven method. On the other hand,  a complexity scalable brightening algorithm is provided by the proposed hybrid learning.  Such a framework is attractive for ``capturing the moment" via mobile computational photography in the coming 5G era. The data-driven method can be adopted to produce an image for previewing on the mobile device. The captured image will be simultaneously sent to the cloud and a high-quality image is produced immediately. The generated image in the cloud will be sent back to the mobile device instantly due to the low latency of the 5G. If the photographer does not like  the synthesized image, she/he can capture another image immediately.

\subsection{Ablation Study}

Since the main objective of this paper is to explore the feasibility of hybrid learning framework rather than a more sophisticated neural network for deep learning, simple ablation study is conducted on the network structure and loss functions.

The ablation study on the BN is shown in Fig. \ref{Fig_f}. Clearly, the BN is helpful for the convergence of the proposed network and it is thus adopted by the proposed network.

\begin{figure}[t]
	\begin{center}
		\includegraphics[width=0.9\linewidth]{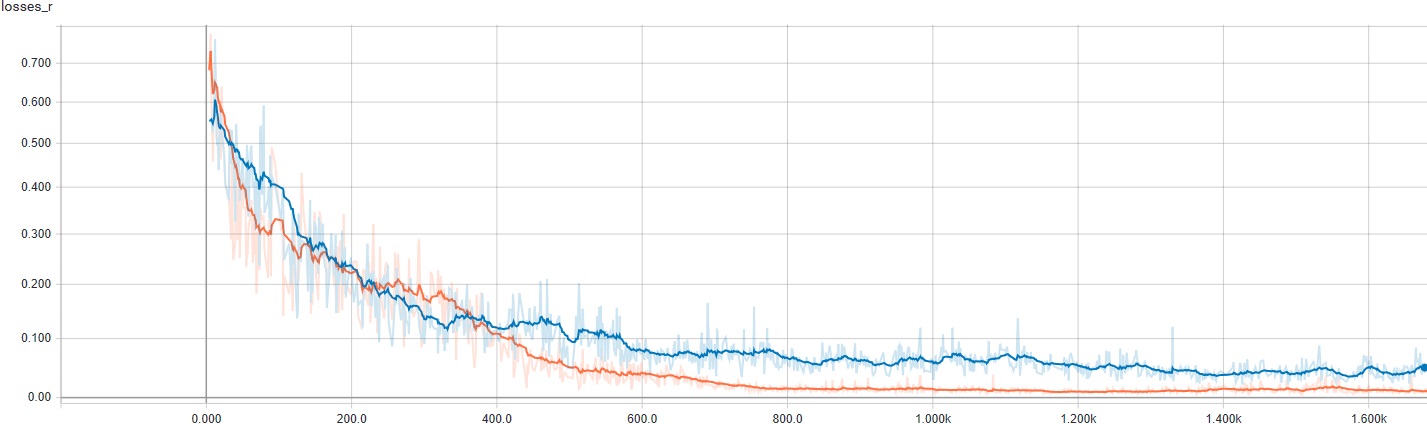}
	\end{center}
	\caption{ Ablation study on loss functions. The Y-axis represents the loss and the X-axis represents the number of iterations. The red line indicates the results with the weights, and the blue line is the results without the weights. Clearly, the adaptive weights can improve the convergence.}
	\label{Fig_five}
\end{figure}

The under-exposed regions contain random noise, and it is difficult to obtain its regular pattern. Although deep learning methods have stronger representation capability, it is still difficult to characterize random noise which brings trouble to network training. Therefore, an adaptive weight is proposed to the restoration loss function as in  the equation (\ref{eq15}), which reduces the influence of noise to the network training. As shown in Fig. \ref{Fig_five},  the change of loss is significantly reduced by adding the weights, which can simplify the network training.

Although the restoration loss $L_r$ can implicitly measure the color difference, it cannot guarantee that $({f_i}(Z_{1})+Z_i)$ and $Z_{Ti}$ have the same color direction. There may exist color distortion by using the restoration loss only, as shown in Fig. \ref{Fig_five1}. By adding the color loss $L_c$, the color distortion can be reduced as indicated by the results in Table  \ref{tab1}.

\subsection{Comparison of Seven Different Brightening Algorithms}

In this subsection,  the proposed algorithm is compared with six state-of-the-art image brightening algorithms including LIME \cite{LIME}, NPE \cite{NPE}, LECARM \cite{LIME}, SNIE \cite{SNIE}, RetinexNet \cite{Retinex-Net} and DeepUPE \cite{DeepUPE}. Both the RetinexNet and the DeepUPE are deep learning methods, and all the others are model-driven methods. The MEF-SSIM in \cite{MEFSSIM} is adopted to objectively assess the performances of the seven algorithms.  The performance assessment is shown in Table \ref{tab2}. Clearly, our algorithm significantly outperforms the other six state-of-art algorithms in terms of the MEF-SSIM evaluation.

\begin{figure*}[htb]
	\centering
	\includegraphics[width=0.9\textwidth]{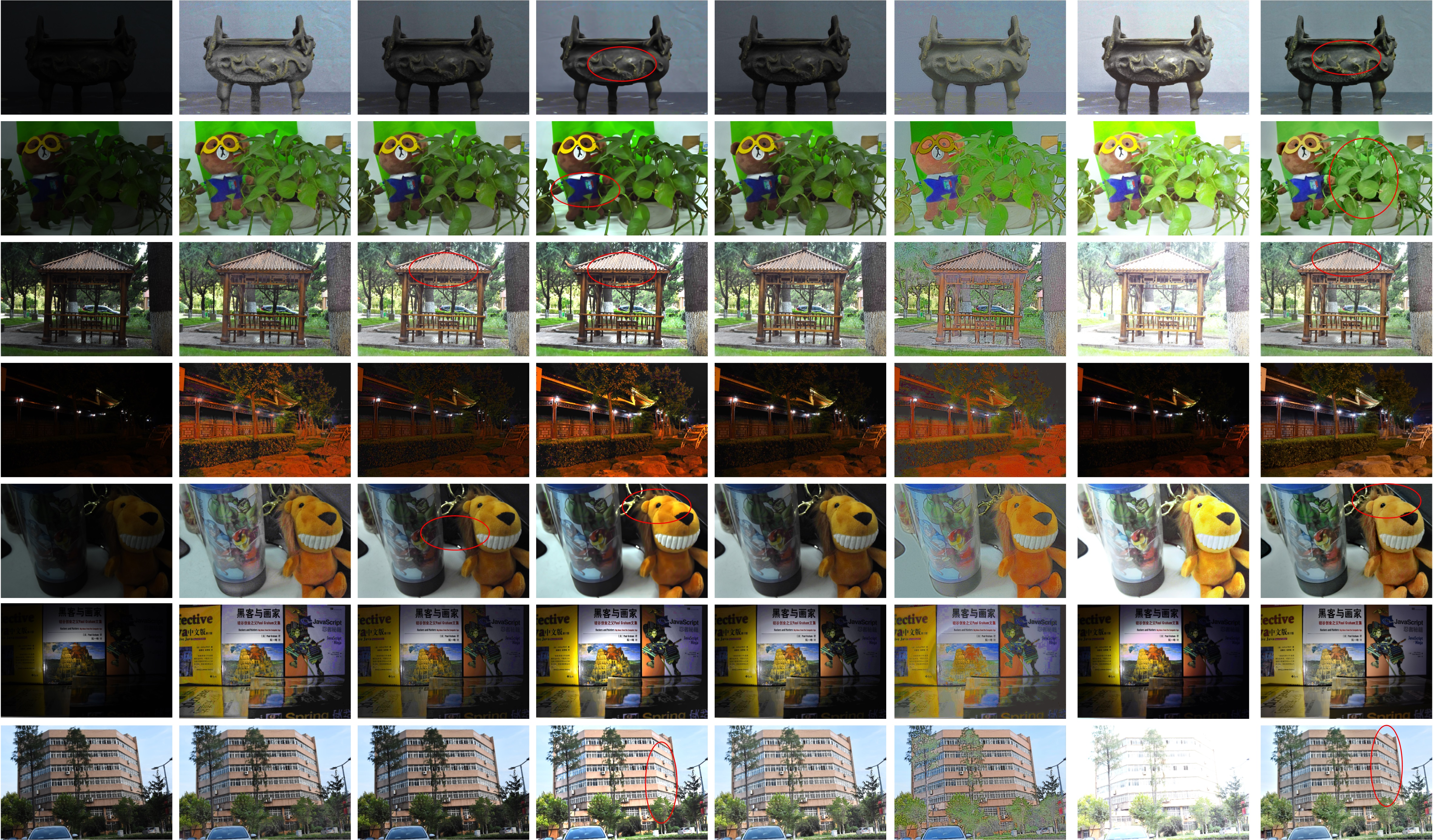}
	\caption{The first column shows the low-lighting images, the second column to the seven columns illustrate the results of NPE, SNIE, LIME, LECARM, RetinexNet and DeepUPE, respectively. The last column shows our results. All the LIME, DeepUPE  and our algorithm can brighten the low-lighting images while reducing noise, but the LIME and DeepUPE sometimes cause parts of the brightened images being washed out.}
	\label{Fig_seven}
\end{figure*}

Besides the objective evaluation, seven brightened images are shown in Fig. \ref{Fig_seven} to demonstrate the effectiveness of the seven algorithms. It can be shown that there are visible distortions in the brightened images by the NPE and RetinexNet. The visibility of brightened images by the SNIE and LECARM needs to be improved. The LIME can achieve good results, but it cannot preserve the details in dark regions. The DeepUPE can achieve the desired effect for dark images, but this method results in over-saturation for bright regions. The proposed algorithm can preserve details, reduce noise, and avoid color distortion. The input image is brightened while the brightest regions are prevented from being washed out.

\begin{table*}[htb]
	\begin{center}
		\centering
		\caption{MEF-SSIM of Seven Different Algorithms}
		\tabcolsep8pt\begin{tabular}{|c|c|c|c|c|c|c|c|}\hline
			image & NPE & SNIE & LIME & LECARM & RetinexNet & DeepUPE & Proposed\\\hline
			furnace & 0.8638 & 0.7923 & 0.8877 & 0.7966 &  0.7787 & 0.9074 & \textbf{0.9218}\\
			tea & 0.8286 & 0.7926 & 0.8788 & 0.8346 &  0.6974 &  0.9069 & \textbf{0.9256}\\
			grass & 0.9134 & 0.8573 & 0.9252 &  0.8976 & 0.7572 & 0.9620 & \textbf{0.9694}\\
			pavilion & 0.8960 & 0.8831 & 0.9272 & 0.9558 & 0.6503 & 0.8062 & \textbf{0.9672}\\
			flower & 0.9060 & 0.8966 & 0.9310 & 0.9578 & 0.7102 & 0.6921 & \textbf{0.9665}\\
			cup &  0.8221 & 0.7639 & 0.8625 & 0.8067 & 0.7278 &  0.8789 & \textbf{0.8965}\\
			flower pot & 0.8107 & 0.6685 & 0.8179 & 0.7106 & 0.6922 & \textbf{0.8879} & 0.8660\\
			succulents & 0.8494 & 0.7744 & 0.8747 & 0.8295 &  0.6812 & 0.8931 & \textbf{0.9163}\\
			electric fan & 0.8826 &  0.8280 & 0.9320 & 0.9098 & 0.7421 &  0.9437 & \textbf{0.9668}\\
			bag &  0.8752 & 0.8429 &  0.9090 & 0.9118 & 0.6622 &  0.9367 & \textbf{0.9701}\\
			dog & 0.8408 & 0.7703 & 0.8620 & 0.8344 & 0.7264 & 0.9247 & \textbf{0.9340}\\
			building & 0.8922 & 0.8847 & 0.9192 & 0.9379 & 0.7401 &  0.6252 & \textbf{0.9411}\\\hline
			average & 0.8651 & 0.8129 & 0.8939 & 0.8653 & 0.7138 & 0.8637 &\textbf{0.9368}\\\hline
		\end{tabular}
		\label{tab2}
	\end{center}
\end{table*}

\begin{figure}[t]
	\begin{center}
	\includegraphics[width=0.9\linewidth]{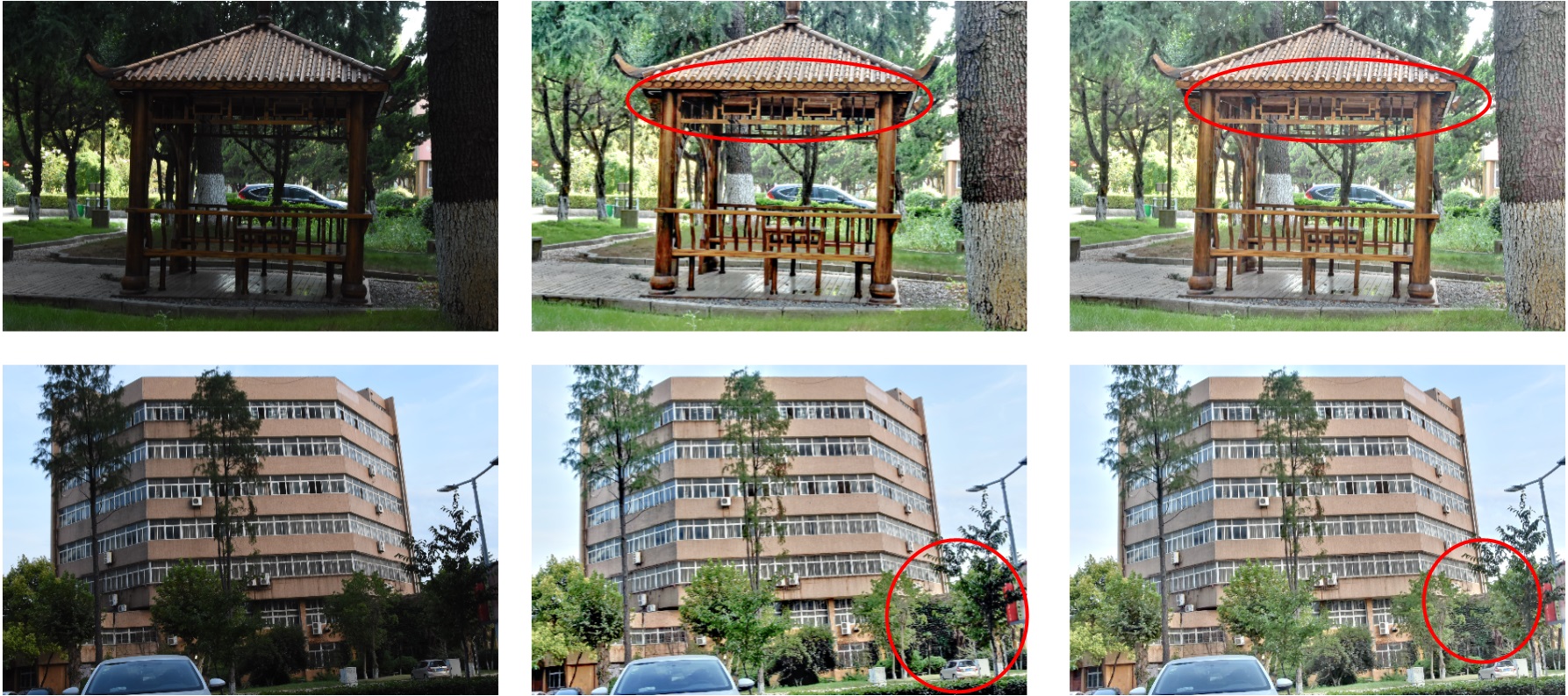}
	\end{center}
	\caption{Comparison of the brightened images by using inaccurate CRF and accurate CRF. The first column shows the input images, the second column illustrates the results by using inaccurate CRF, and the third column demonstrates brightened images by using accurate CRF.}
    \label{Fig_nine}
\end{figure}

\subsection{Limitation of the Proposed Algorithm}
Although the proposed algorithm outperforms the other six algorithms, there is space for further study. For example, the proposed algorithm assumes  that the accurate CRFs are available. For images from unknown sources, it is quite challenging to estimate the  CRFs. If the estimated CRFs are not accurate,  the difference between the virtual images and their ground truth images become large, and the quality of the brightened images drops a little bit, as shown in Fig. \ref{Fig_nine}.

As indicated in \cite{CRFs}, the CRFs can be estimated correctly if the lighting condition is not changed for all the differently exposed images. This is also required by the proposed algorithm to generate the two virtual images even though the MEF is not sensitive to the variable lighting environment \cite{Mertens09}.

\section{CONCLUSION REMARKS}
\label{paradigm5}
A new hybrid learning framework is introduced to study single image brightening. This paper was focuses on the compensation of  model-driven method and data-driven method rather than employing sophisticated neural networks. Two initial virtual images with large exposure times are first generated by using the model-driven method and they are enhanced by using the data-driven residual convolutional neural networks. All the input image and the two virtual images are fused to obtain a brightened image. The brightness of the input image is increased while the high-light regions are prevented from being washed out. Experimental results show that the proposed algorithm outperforms the existing algorithms.

It was assumed by the proposed algorithm that the camera response functions (CRFs) are available. This is not an issue if the proposed algorithm is embedded into a digital camera or a smart phone but it might not be true for an image downloaded from the Internet. A deep learning based algorithm could be used or a single image based estimation method such as the LECARM \cite{LIME} is adopted to estimate the CRFs. Two virtual images with 2EV gaps are generated in the proposed algorithm. It is interesting but challenging to determine the optimal number of virtual images with the optimal EV gaps. One more interesting topic is to apply the proposed hybrid learning for other image processing problems. It is also interesting to develop more sophisticated deep learning methods to replace the one used in this paper. All of them will be studied in our future R$\&$D.

\section*{acknowledgement}
This work was supported by the National Nature Science Foundation of China under Project 61775172 and Project 61620106012.

\end{document}